\documentclass[11pt, letterpaper]{article}
\usepackage{graphicx}
\usepackage[caption=false]{subfig}
\usepackage[utf8]{inputenc}
\usepackage[margin=1.5cm]{geometry}
\usepackage{titlesec}
\usepackage{tabu}
\usepackage{enumitem}
\usepackage{longtable}
\usepackage{amssymb}
\usepackage{amsmath}
\usepackage{xcolor}
\newlist{selectlist}{itemize}{2}
\setlist[selectlist]{label=$\square$,leftmargin=*,noitemsep,topsep=0pt}

\usepackage{lmodern}

\usepackage{hyperref}
\hypersetup{
    colorlinks=true,
    linkcolor=blue,
    filecolor=magenta,      
    urlcolor=blue,
}
 
\titleformat{\section}[block]{\hspace{1em}\bfseries}{\thesection.}{0.5em}{} 
\titleformat{\subsection}[block]{\hspace{1em}}{\thesubsection}{0.5em}{}

\begin{document}

\begin{flushleft}

\setlength{\parindent}{0pt}
\setlength{\parskip}{10pt}

\textbf{Article title}\\ OpenFish: Biomimetic Design of a Soft Robotic Fish for High Speed Locomotion

\textbf{Authors}\\ Sander C.~van den Berg$^{1,\dagger}$, Rob B.N.~Scharff$^{*2,\dagger}$, Zolt{\'a}n Rus{\'a}k$^{1}$, and Jun~Wu$^{*1}$

\textbf{Affiliations}\\ {$^{1}$Department of Sustainable Design Engineering, Delft University of Technology, \\ Landbergstraat 15, 2628 CE, Delft, The Netherlands\\
$^{2}$ Bioinspired Soft Robotics Lab, Istituto Italiano di Tecnologia (IIT), \\ Via Morego 30, 16163, Genova, Italy}\\
$^{\dagger}$ Contributed equally to this manuscript

\textbf{Corresponding author’s email address and Twitter handle}\\
Rob Scharff - rob.scharff@iit.it (@Rob\_Scharff)\\
Jun Wu - j.wu-1@tudelft.nl

\textbf{Abstract}\\ 
We present OpenFish: an open source soft robotic fish which is optimized for speed and efficiency. The soft robotic fish uses a combination of an active and passive tail segment to accurately mimic the thunniform swimming mode. Through the implementation of a novel propulsion system that is capable of achieving higher oscillation frequencies with a more sinusoidal waveform, the open source soft robotic fish achieves a top speed of $0.85~\mathrm{m/s}$. Hereby, it outperforms the previously reported fastest soft robotic fish by $27\%$. Besides the propulsion system, the optimization of the fish morphology played a crucial role in achieving this speed. In this work, a detailed description of the design, construction and customization of the soft robotic fish is presented. Hereby, we hope this open source design will accelerate future research and developments in soft robotic fish.


\newpage
\textbf{Specifications table}\\
\vskip 0.2cm
\tabulinesep=1ex
\begin{tabu} to \linewidth {|X|X[3,l]|}
\hline  \textbf{Hardware name} & OpenFish - Open source soft robotic fish
  \\
  \hline \textbf{Subject area} & 
  \begin{itemize}[noitemsep, topsep=0pt]
  \item Soft robotics
  \item Marine robotics
  \item Oscillating propulsion
  \end{itemize}
  \\
  \hline \textbf{Hardware type} &
  \begin{itemize}[noitemsep, topsep=0pt]
  \item Underwater exploration
  \end{itemize}
  \\ 
\hline \textbf{Closest commercial analog} & No commercial analog is available
  \\
\hline \textbf{Open source license} & CC-BY-SA 4.0
  \\
\hline \textbf{Cost of hardware} &
€~160,32
  \\
\hline \textbf{Source file repository} & 
 
      \href{https://doi.org/10.17605/OSF.IO/FB5VH}{https://doi.org/10.17605/OSF.IO/FB5VH}
  
\\\hline
\end{tabu}
\end{flushleft}

\newpage

\section{Hardware in context}
\label{sec:Hardware_in_context}
Oscillating propulsion for underwater locomotion offers several potential advantages over rotary propulsion. First, the high operating frequencies of rotary propulsion systems create disturbing vibrations in the water. In contrast, oscillating propulsion systems require much lower operating frequencies than rotary propulsion systems and consequently do not create such disturbing vibrations. Second, rotary propulsion systems actively suck in objects and wildlife into the propeller, whereas systems with oscillating propulsion pushes obstacles away from its body rather than entangling them. Third, the oscillating motion can outperform rotary propulsion in terms of efficiency. Lab experiments have demonstrated efficiencies up to $87\%$~\cite{anderson1998oscillating}, whereas rotary systems with small propellers typically do not reach efficiencies above $40\%$~\cite{triantafyllou1995efficient}. Rotary propulsion suffers from energy losses resulting from the rotating water flow. Moreover, a pressure needs to be exerted on the rotary shaft to prevent water from seeping into the system, resulting in energy loss due to friction. The required pressure to be applied on the shaft, and thus loss of energy, gets especially large when the underwater system is operating at large depths. Oscillating propulsion systems do not have rotating parts in direct contact with the water and therefore do not suffer from this problem. On top of that, oscillating propulsion systems harvest energy from the turbulence at the wake of the vessel's body. The abovementioned advantages have motivated the development of robotic fish with oscillating propulsion for applications such as trait-based ecology investigation~\cite{Romano2019}, studying the role of social robots in treating fish anxiety~\cite{Romano2021}, studying collective fish behavior~\cite{Romano2022}, studying the design principles and coordination mechanisms of biological systems~\cite{Lee2022}, energy harvesting~\cite{Tan2021}, tracking of real fish or underwater vehicles~\cite{Yu2021}, science education~\cite{Phamduy2015}, and water quality monitoring~\cite{Kai2020}.

\begin{figure}[h!]
\centering
\includegraphics[width=.8\textwidth]{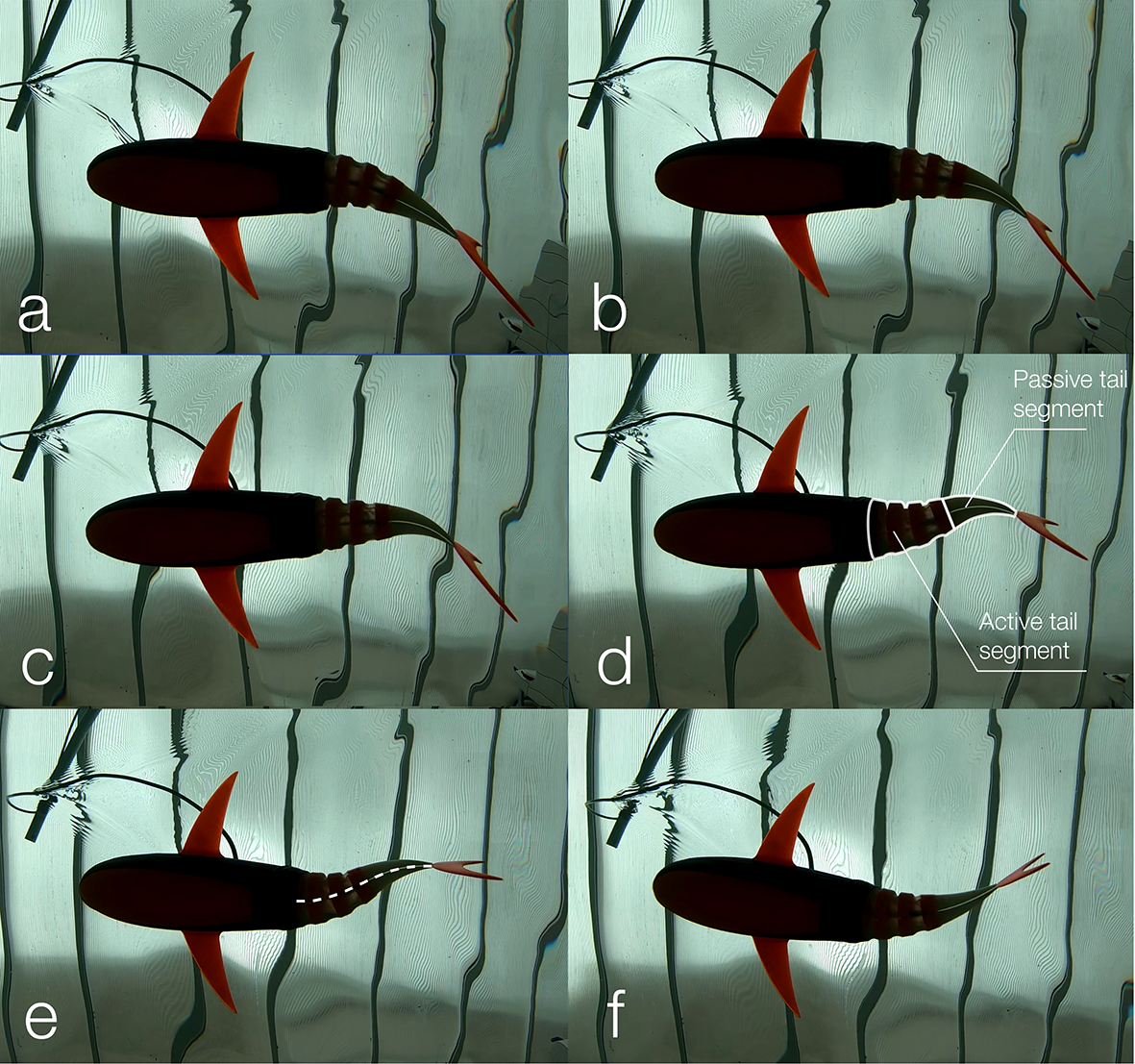}
\caption{The swimming pattern of the soft robotic fish presented in this work in chronologically ordered snapshots from a to f. The thunniform-like swimming mode is realized through a combination of an active and a passive tail segment.}
\label{fig:fishlocomotion}
\end{figure}

Despite their advantages and applications, the study and development of robotic fish for underwater locomotion is progressing slowly. We argue that one of the main reasons for this slow progress is the lack of a suitable open source design that can function as a shared platform for research and development. Building a robotic fish with satisfactory performance from scratch is a time-consuming process that typically requires extensive tuning of design parameters based on literature research and experiments over the course of many design iterations. Moreover, it is difficult to compare results that were obtained on the strongly deviating designs that have been previously developed. For example, the size of a robotic fish will greatly influence the evaluation of its swimming speed (even when comparing speed in body lengths per second ($\mathrm{BL/s}$)), and a robotic fish that is tested with a fixed head in a laminar flow tank~\cite{Zhu2019} is likely to generate more thrust than a freely swimming robotic fish with a swaying head~\cite{Katzschmann2018}.

A suitable platform that allows for further investigation of all of the above-mentioned applications of robotic fish requires a versatile design that is capable of achieving high speeds as well as safe and compliant behavior. Moreover, the fabrication of the basic design should be accessible, fast, and low-cost. Finally, it should be easy to modify the design and to build in additional functionality. Most robotic fish do not meet all of the above-mentioned criteria. For example, robotic fish designs that are capable of achieving speeds of $1.02~\mathrm{m/s}$~\cite{Zhu2019} or even $2.0~\mathrm{m/s}$~\cite{White2021} make use of a rotary shaft propulsion mechanism and therefore lack compliance and tunability. On the other hand, designs that make use of a compliant tail (i.e. soft robotic fish) are typically not capable of high speed swimming~\cite{Katzschmann2018,mazumdar2008maneuverability}. Moreover, the fabrication of most of these existing designs is complex and tedious.

In our previous work, we presented a soft robotic fish with a compliant tail which was optimized for speed and efficiency~\cite{vandenBerg2020} (see Fig.~\ref{fig:fishlocomotion}). A key innovation of our system is the use of the continuous rotation of a DC motor to pull the cables connected to both sides of the active tail segment, instead of the commonly used back-and-forth motion of a servo-motor. As a result, higher frequencies can be obtained with a more sinusoidal waveform. Our fish is able to reach speeds up to $0.85$~m/s. Hereby, it outperforms the previously reported fast soft robotic fish by Zhong et al.~\cite{zhong2017novel} with a significant margin of $27\%$. Moreover, the fabrication of the design is accessible, low-cost, and easy to modify. Hereby, it makes for an excellent platform for future research and development.

The rationale behind the design of the oscillating propulsion system and the fish morphology has been discussed extensively in our previous work and is based on literature and modeling~\cite{vandenBerg2020}. Instead, this work will focus on presenting a detailed description of the open source OpenFish design that will allow others to replicate and build upon this design. All OpenFish design files are freely accessible at \href{https://doi.org/10.17605/OSF.IO/FB5VH}{https://doi.org/10.17605/OSF.IO/FB5VH}.

\section{Hardware description}
\label{sec:Hardware_description}

\begin{figure}[h]
\centering%
\includegraphics[width=\textwidth]{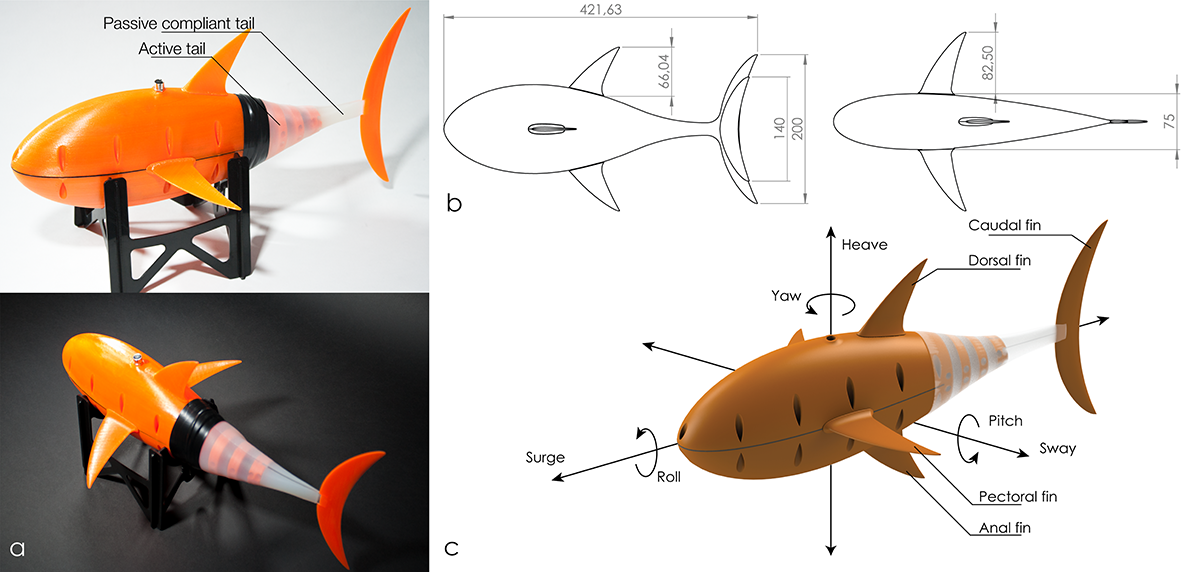}
\caption{(a) Design of the open source soft robotic fish, (b) Side and top view of the soft robotic fish with its main dimensions, (c) explanation of the terms used for fish anatomy and fish stability.}

\label{fig:design}
\end{figure}

The biomimetic design of the open source soft robotic fish is shown in Fig.~\ref{fig:design}. The design has been optimized for speed and efficiency. A single DC motor is used to realize the oscillating motion for propulsion. A gearbox converts the continuous rotation of the motor to alternately pull the cables on both sides of the active tail segment. Hereby, the system is capable of achieving much higher oscillation frequencies with a more sinusoidal waveform than systems that use the back-and-forth motion of a servo-motor to pull the cables. The passive compliant tail segment then conforms to the environment to accurately mimic the thunniform swimming mode. The shape of the body and fins have also been optimized for speed and efficiency. For a detailed description of the rationale behind the design of the oscillating propulsion system and the fish morphology we refer the reader to our previous work~\cite{vandenBerg2020}.
Instead, this section will focus on explaining how OpenFish's versatile design can be deployed for a wide range of applications. Specifically, we envision the OpenFish design to aid in verifying and comparing research on oscillating propulsion, to aid the development of a fully autonomous and intelligent soft robotic fish, and to aid studies that require underwater task performance and data collection. Finally, subsection~\ref{sec:sec:usecase} presents a use case to demonstrate OpenFish's well-suited behavior for verifying a theoretical model that describes the influence of the caudal fin size on the swimming speed. 

\begin{enumerate}
    \item \textbf{Verification and comparison of research on oscillating propulsion} \\
    The OpenFish design can be used to verify modeling results on oscillating propulsion and investigation of design parameters by modifying the OpenFish design and comparing it to the benchmark design. In combination with the standard test protocol described in Section~\ref{sec:Operation_instructions}, the OpenFish design can act as a platform for fair comparisons of fish performance. Design parameters that might be of interest include the lengths of the active and passive tail, the stiffness of the compliant tail, the shape of the head, and the shape or placement of the fins. As the OpenFish design is mostly composed of lasercut and 3D-printed parts, the morphology and materials of its body can be easily modified. At the same time, the low costs, low number of parts, and ease of assembly and control of the OpenFish make construction of the fish for the purpose of verification and comparison accessible.

    \item \textbf{Development of a fully autonomous and intelligent soft robotic fish} \\
    The OpenFish design is expected to support the development of a fully autonomous and intelligent soft robotic fish with increased functionality. The OpenFish design is built in anticipation of such additions. For example, the current propulsion system allows for an easy implementation of a steering mechanism, through rotation of the attachment plate that holds the spur gears around the main motor axis. A bearing (\textbf{Part \#34}) has already been included in the design for this purpose and the placement of the gear for steering (\textbf{Part \#29}) is indicated in \textbf{Assembly \#1}. Other envisioned developments are the inclusion of a proprioceptive sensing system in the tail and the inclusion of a method to change the stiffness of the passive compliant tail segment dynamically. Through these additions, the swimming mode of the fish could be optimized at different tail beat frequencies. Finally, the design could be further enhanced by replacing the cable that is attached to the fish by an embedded power supply and a wireless communication system or autonomous control system. The head of the OpenFish design leaves plenty of space to accommodate such additions.
    
    \item \textbf{Underwater exploration and task performance}\\
    Lastly, the OpenFish design is expected to be deployed for exploring, monitoring, and performing tasks in an underwater environment. Due to using oscillating propulsion and a compliant tail, OpenFish can operate without disturbing or damaging underwater flora and fauna. Its ability to blend in with its environment makes it a valuable tool to study the behavior of underwater animals. The design can be easily enriched with exteroceptive sensors (e.g. cameras and chemical sensors) and grippers to collect the required data.
\end{enumerate}

\subsection{Use case: Verification of predicted swimming speed based on scaling of the caudal fin}
\label{sec:sec:usecase}

The caudal fin is of great importance to the swimming characteristics of fish and has therefore been extensively studied. However, it has been difficult to compare the performance of different caudal fin designs on a robotic fish due to the large variety of robotic fish embodiments that are used to test their performance.

This section will highlight the use of the OpenFish platform to verify a simple model that predicts the relative speed of the robotic fish based on the scaling factor of the caudal fin.

Experiments were performed with both a small caudal fin (height of $140~\mathrm{mm}$, surface area $A_\mathrm{s} = 2431~\mathrm{mm}^2$) and a large caudal (height of $200~\mathrm{mm}$, surface area $A_\mathrm{l} = 4065~\mathrm{mm}^2$) (see Fig.~\ref{fig:design}(c)).

The effect of an increased caudal fin surface area on the swimming speed can be estimated as follows. The hydrodynamic drag can be expressed as:
\begin{equation}
    F_\mathrm{D}=\frac{1}{2}~\rho~S~v^2~C_\mathrm{D}
    \label{eq:drag}
\end{equation}
Where $\rho$ is the density of the fluid, $v$ is the speed, $S$ is the reference area, and $C_\mathrm{D}$ is the drag coefficient.
When swimming at a constant speed, the drag force and the thrust force ($F_\mathrm{t}$) are in equilibrium ($F_\mathrm{D} = F_\mathrm{t}$). As the density of the fluid and drag coefficient are constant over the two experiments, it follows that:
\begin{equation}
    F_\mathrm{t}=v^2~\mathrm{c}
\end{equation}
Where $\mathrm{c}$ is a constant replacing all constants from equation~\ref{eq:drag}. As the thrust force increases linearly with the surface area of the caudal fins, the expected swimming speed of the fish with the large caudal fin ($v_\mathrm{l}$) can be expressed as a function of the swimming speed of the fish with the smaller caudal fin ($v_\mathrm{s}$) and the surface areas of both caudal fins:
\begin{equation}
    v_\mathrm{l}=\sqrt{\frac{A_\mathrm{l}~ v_\mathrm{s}^2}{A_\mathrm{s}}}
    \label{eq:predictedspeed}
\end{equation}
It should be noted that changing the size and shape of the caudal fin is likely to lead to changes in the angle of attack. This effect is not taken into account for the estimation above.

\vskip 0.3cm \noindent
Features:
\begin{itemize}

    \item Affordable and easy to fabricate design for verification and comparison of research on oscillating propulsion
    \item Platform for the development of a fully autonomous and intelligent soft robotic fish: The design allows for easy implementation of additional features such as a steering mechanism and (proprioceptive) sensors
    \item High-speed and efficient free-swimming robotic fish with a compliant tail for underwater exploration and task performance
\end{itemize}

\section{Design files summary}
\label{sec:Design_files_summary}

\vskip 0.1cm
\tabulinesep=1ex
\begin{tabu} to \linewidth {|X|X[1.5,1]|X|X[1.5,1]|} 
\hline
\textbf{Design filename} & \textbf{File type} & \textbf{Open source license} & \textbf{Location of the file} \\\hline

Part \#1 & STL file; CAD file (SLDPRT) & CC & Available with the article \\\hline
Part \#2 & STL file; CAD file (SLDPRT)& CC & Available with the article \\\hline
Part \#3 & STL file; CAD file (SLDPRT)& CC & Available with the article \\\hline
Part \#4 & STL file; CAD file (SLDPRT)& CC & Available with the article \\\hline
Part \#5 & STL file; CAD file (SLDPRT)& CC & Available with the article \\\hline
Part \#6 & STL file; CAD file (SLDPRT)& CC & Available with the article \\\hline
Part \#7 & STL file; CAD file (SLDPRT)& CC & Available with the article \\\hline
Part \#8 & STL file; CAD file (SLDPRT)& CC & Available with the article \\\hline
Part \#9 & STL file; CAD file (SLDPRT)& CC & Available with the article \\\hline
Part \#10 & CAD file (SLDPRT)& CC & Available with the article \\\hline
Part \#11 & STL file; CAD file (SLDPRT)& CC & Available with the article \\\hline
Part \#12 & STL file; CAD file (SLDPRT)& CC & Available with the article \\\hline
Part \#13 & DXF file; CAD file (SLDPRT)& CC & Available with the article \\\hline
Part \#14 & DXF file; CAD file (SLDPRT)& CC & Available with the article \\\hline
Part \#15 & STL file; CAD file (SLDPRT)& CC & Available with the article \\\hline
Part \#16 & STL file; CAD file (SLDPRT)& CC & Available with the article \\\hline
Part \#17 & STL file; CAD file (SLDPRT)& CC & Available with the article \\\hline
Part \#18 & DXF file; CAD file (SLDPRT)& CC & Available with the article \\\hline
Part \#19 & STL file; CAD file (SLDPRT)& CC & Available with the article \\\hline
Part \#20 & DXF file; CAD file (SLDPRT)& CC & Available with the article \\\hline
Part \#21 & DXF file; CAD file (SLDPRT)& CC & Available with the article \\\hline
Part \#22 & DXF file; CAD file (SLDPRT)& CC & Available with the article \\\hline
Part \#23 & DXF file; CAD file (SLDPRT)& CC & Available with the article \\\hline
Part \#24 & DXF file; CAD file (SLDPRT)& CC & Available with the article \\\hline
Part \#25 & DXF file; CAD file (SLDPRT)& CC & Available with the article \\\hline
Assembly \#1 & CAD file (SLDASM, Parasolid)& CC & Available with the article \\\hline
Assembly \#2 & CAD file (SLDASM, Parasolid)& CC & Available with the article \\\hline

\end{tabu}

\vskip 0.3cm
\noindent
\textbf{Part \#1} is the bottom part of the fish head\\
\textbf{Part \#2} is the top part of the fish head\\
\textbf{Part \#3} is the large caudal fin of the fish\\
\textbf{Part \#4} is both the left and right pectoral fin of the fish \\
\textbf{Part \#5} is both the anal and dorsal fin of the fish \\
\textbf{Part \#6} is the first rib of the active tail segment\\
\textbf{Part \#7} is the second rib of the active tail segment\\
\textbf{Part \#8} is the third rib of the active tail segment\\
\textbf{Part \#9} is the fourth and last rib of the active tail segment\\
\textbf{Part \#10} is the silicone tail cover\\
\textbf{Part \#11} is both the left and right part of the outer mold used to cast \textbf{Part \#10}\\
\textbf{Part \#12} is the inner mold used to cast \textbf{Part \#10}\\
\textbf{Part \#13} is an insert to the inner mold used to cast \textbf{Part \#10}\\
\textbf{Part \#14} is the rubber sheet that seals \textbf{Part \#1} and \textbf{Part \#2} \\
\textbf{Part \#15} is the mount for the motor assembly\\
\textbf{Part \#16} is the mount for the bearing\\
\textbf{Part \#17} is the mount for the gearbox assembly\\
\textbf{Part \#18} is the compliant tail \\
\textbf{Part \#19} is the small caudal fin of the fish\\
\textbf{Part \#20} is a plate in the gearbox assembly\\
\textbf{Part \#21} is a mounting plate of the gearbox assembly\\
\textbf{Part \#22} is the pushrod connector plate\\
\textbf{Part \#23} is the front plate of the fish mount\\
\textbf{Part \#24} is the back plate of the fish mount\\
\textbf{Part \#25} is the side plate of the fish mount\\
\textbf{Assembly \#1} is the assembly of the soft robotic fish that consists of all of the above mentioned parts (besides \textbf{Part \#11}, \textbf{Part \#12}, and \textbf{Part \#13}) as well as the off-the-shelf parts \\
\textbf{Assembly \#2} is the assembly of the mold for part \textbf{Part \#10} \\

\section{Bill of materials summary}
\label{sec:Bill_of_materials_summary}
\vskip 0.2cm
\tabulinesep=1ex
\noindent
\begin{longtabu} to \linewidth {|X[1.1,1]|X|X[0.6,1]|X[0.8,1]|X|X|X[1.1,1]|}
\hline
\textbf{Designator} & \textbf{Component} & \textbf{Number} & \textbf{Cost per unit - currency} & \textbf{Total cost - currency} & \textbf{Source of materials} & \textbf{Material type} \\\hline
Part \#1 &  & 1 & €~19,50/kg & €~4,88 (250g) & \href{https://www.123-3d.nl/123-3D-Filament-oranje-2-85-mm-PLA-1-kg-Jupiter-serie-i1843-t7346.html}{123-3D} & PLA \\\hline
Part \#2 &  & 1 & €~19,50/kg & €~2,91 (149g) & \href{https://www.123-3d.nl/123-3D-Filament-oranje-2-85-mm-PLA-1-kg-Jupiter-serie-i1843-t7346.html}{123-3D} & PLA \\\hline
Part \#3 &  & 1 & €~19,50/kg & €~0,31 (16g) & \href{https://www.123-3d.nl/123-3D-Filament-oranje-2-85-mm-PLA-1-kg-Jupiter-serie-i1843-t7346.html}{123-3D} & PLA \\\hline
Part \#4 &  & 2 & €~19,50/kg & €~0,39 (2x10g) & \href{https://www.123-3d.nl/123-3D-Filament-oranje-2-85-mm-PLA-1-kg-Jupiter-serie-i1843-t7346.html}{123-3D} & PLA \\\hline
Part \#5 &  & 2 & €~19,50/kg & €~0,35 (2x9g) & \href{https://www.123-3d.nl/123-3D-Filament-oranje-2-85-mm-PLA-1-kg-Jupiter-serie-i1843-t7346.html}{123-3D} & PLA \\\hline
Part \#6 &  & 1 & €~19,50/kg & €~0,39 (20g) & \href{https://www.123-3d.nl/123-3D-Filament-oranje-2-85-mm-PLA-1-kg-Jupiter-serie-i1843-t7346.html}{123-3D} & PLA \\\hline
Part \#7 &  & 1 & €~19,50/kg & €~0,18 (9g) & \href{https://www.123-3d.nl/123-3D-Filament-oranje-2-85-mm-PLA-1-kg-Jupiter-serie-i1843-t7346.html}{123-3D} & PLA \\\hline
Part \#8 &  & 1 & €~19,50/kg & €~0,14 (7g) & \href{https://www.123-3d.nl/123-3D-Filament-oranje-2-85-mm-PLA-1-kg-Jupiter-serie-i1843-t7346.html}{123-3D} & PLA \\\hline
Part \#9 &  & 1 & €~19,50/kg & €~0,14 (7g) & \href{https://www.123-3d.nl/123-3D-Filament-oranje-2-85-mm-PLA-1-kg-Jupiter-serie-i1843-t7346.html}{123-3D} & PLA \\\hline
Part \#10 &  & 1 & €~36,41/pot (0.91~kg) & €~2,00 (50g) & \href{https://www.formx.eu/molding--casting/platinum-silicones/mold-star--series/mold-star15-slow----091-kg.php}{FormX} & Silicone - Shore 15\\\hline
Part \#11 &  & 2 & €~19,50/kg & €~6,24 (2x160g) & \href{https://www.123-3d.nl/123-3D-Filament-oranje-2-85-mm-PLA-1-kg-Jupiter-serie-i1843-t7346.html}{123-3D} & PLA \\\hline
Part \#12 &  & 1 & €~19,50/kg & €~3,41 (175g) & \href{https://www.123-3d.nl/123-3D-Filament-oranje-2-85-mm-PLA-1-kg-Jupiter-serie-i1843-t7346.html}{123-3D} & PLA \\\hline
Part \#13 & 1~mm thickness & 1 & €~$25,95/\mathrm{m}^2$ & €~0,03 ($1209.16~\mathrm{mm}^2$) & \href{https://www.tudelft.nl/io/over-io/faciliteiten/pmb/materialenshop}{TU Delft} & PETG \\\hline
Part \#14 & 1~mm thickness & 1 & €~$18,26/\mathrm{m}^2$ & €~0,05 ($2981.87~\mathrm{mm}^2$) & \href{https://www.tudelft.nl/io/over-io/faciliteiten/pmb/materialenshop}{TU Delft} & Nitril Rubber \\\hline
Part \#15 &  & 1 & €~19,50/kg & €~0,35 (18g) & \href{https://www.123-3d.nl/123-3D-Filament-oranje-2-85-mm-PLA-1-kg-Jupiter-serie-i1843-t7346.html}{123-3D} & PLA \\\hline
Part \#16 &  & 1 & €~19,50/kg & €~0,04 (2g) & \href{https://www.123-3d.nl/123-3D-Filament-oranje-2-85-mm-PLA-1-kg-Jupiter-serie-i1843-t7346.html}{123-3D} & PLA \\\hline
Part \#17 &  & 1 & €~19,50/kg & €~0,49 (25g) & \href{https://www.123-3d.nl/123-3D-Filament-oranje-2-85-mm-PLA-1-kg-Jupiter-serie-i1843-t7346.html}{123-3D} & PLA \\\hline
Part \#18 & 1~mm thickness & 1 & €~$25,95/\mathrm{m}^2$ & €~0,07 ($2616.66~\mathrm{mm}^2$)  & \href{https://www.tudelft.nl/io/over-io/faciliteiten/pmb/materialenshop}{TU Delft} & PETG \\\hline
Part \#19 &  & 1 & €~19,50/kg & €~0,20 (10g) & \href{https://www.123-3d.nl/123-3D-Filament-oranje-2-85-mm-PLA-1-kg-Jupiter-serie-i1843-t7346.html}{123-3D} & PLA \\\hline
Part \#20 & 2~mm thickness & 1 & €~$133,10/\mathrm{m}^2$ & €~0,01 ($98.96~\mathrm{mm}^2$) & \href{http://www.stoutperspex.nl/kunststof-producten/pom-delrin-plaat-naturel/570}{Stoutperspex} & POM \\\hline
Part \#21 &  2~mm thickness & 1 & €~$133,10/\mathrm{m}^2$ & €~0,17 ($1260.76~\mathrm{mm}^2$)& \href{http://www.stoutperspex.nl/kunststof-producten/pom-delrin-plaat-naturel/570}{Stoutperspex} & POM \\\hline
Part \#22 & 2~mm thickness & 2 & €~$133,10/\mathrm{m}^2$ & €~0,10 (2x $366.78~\mathrm{mm}^2$) & \href{http://www.stoutperspex.nl/kunststof-producten/pom-delrin-plaat-naturel/570}{Stoutperspex} & POM \\\hline
Part \#23 &  2~mm thickness & 1 & €~$133,10/\mathrm{m}^2$ & €~0,86 ($6448.42~\mathrm{mm}^2$)& \href{http://www.stoutperspex.nl/kunststof-producten/pom-delrin-plaat-naturel/570}{Stoutperspex} & POM \\\hline
Part \#24 &  2~mm thickness & 1 & €~$133,10/\mathrm{m}^2$ & €~0,93 ($7002.78~\mathrm{mm}^2$)& \href{http://www.stoutperspex.nl/kunststof-producten/pom-delrin-plaat-naturel/570}{Stoutperspex} & POM \\\hline
Part \#25 &  2~mm thickness & 2 & €~$133,10/\mathrm{m}^2$ & €~2,06 (2 x $7749.66~\mathrm{mm}^2$)& \href{http://www.stoutperspex.nl/kunststof-producten/pom-delrin-plaat-naturel/570}{Stoutperspex} & POM \\\hline
Part \#26 & 9.7:1 Metal Gearmotor 25Dx48L mm HP 12V & 1 & €~19.37/pc & €~19.37 & \href{https://www.exp-tech.de/motoren/dc-getriebemotoren/7579/9.7-1-metal-gearmotor-25dx48l-mm-hp-12v}{Exptech} & Other \\\hline
Part \#27 & 4x4mm motorshaft coupling & 1 & €~6.95/pc & €~6.95 & \href{https://www.robots4all.be/Webwinkel-Product-133090497/Flexibel-koppelstuk-4x4-mm.html}{Robots4all} & Other \\\hline
Part \#28 & Spur gear M0.5 20T & 1 & €~4,35/pc & €~4,35 & \href{https://benl.rs-online.com/web/p/spur-gears/5217023}{RS online} & POM \\\hline
Part \#29 & Spur gear M0.5 50T & 2 & €~5,03/pc & €~10,06 & \href{https://benl.rs-online.com/web/p/spur-gears/5217146/}{RS online} & POM \\\hline
Part \#30 & Gear for steering & 1 & €~4,30 & €~4,30 & \href{https://www.norelem.com/us/en/Products/Product-overview/Systems-and-components-for-machine-and-plant-construction/22000-Drive-technology/Gears-Gear-Racks-Bevel-Gears/22402-Spur-gears-plastic-module-1-injection-moulded-straight-teeth-engagement-angle-20\%C2\%B0.html}{Norelem} & POM \\\hline
Part \#31 & Huco L gearbox, 1:1 gear ratio & 1 & €~39,10/pc & €~39,10 & \href{https://benl.rs-online.com/web/p/gearboxes/0748437/}{RS online} & Other \\\hline
Part \#32 & Flat washer M5 & 6 & €~3,07/50pcs & €~0,37 & \href{https://nl.rs-online.com/web/p/tap-washers/0525751}{RS Online} & Nylon \\\hline
Part \#33 & Spacer bolt & 4 & €~22,60/50pcs & €~1,81 & \href{https://benl.rs-online.com/web/p/standoffs/0221134}{RS online} & Other \\\hline
Part \#34 & Push rod connector & 2 & €~5,45/5pcs & €~2,18 & \href{https://quartel.nl/product/pichlerez-connector-5-pic5813/}{Quartel} & Other \\\hline
Part \#35 & Bearing & 1 & €~3,23/pc & €~3,23 & \href{https://benl.rs-online.com/web/p/ball-bearings/6189979/}{RS online} & Other \\\hline
Part \#36 & Rivet screw & 2 & €~10.89/100pcs & €~0.22& \href{https://www.thesolutionshop.com/tube-m4-x-15-mm-brass-plated.html}{The solution shop} & Other \\\hline
Part \#37 & Adhesive weights 5g & 50 & €~15,13/50pcs & €~15,13 & \href{https://valkenpower.com/werkplaatsuitrusting/bandenservice-215601931/wielgewicht-ijzer/plaklood-ijzer-1421223113/plaklood-12-x-5gr-50st.html}{Valkenpower} & Other \\\hline
Part \#38 & Steel braided fishing line & 2 & €~3,71/2pcs & €~3,71 & \href{https://www.henkkoster.nl/spro-predator-wire-leader-7x7-snap.html}{Henkkoster} & Other \\\hline
Part \#39 & watertight connector male & 1 & €~12,72/pc & €~12,72 & \href{https://benl.rs-online.com/web/p/industrial-circular-connectors/1748738/?relevancy-data=7365617263685F636173636164655F6F726465723D31267365617263685F696E746572666163655F6E616D653D4931384E525353746F636B4E756D626572267365617263685F6C616E67756167655F757365643D656E267365617263685F6D617463685F6D6F64653D6D61746368616C6C267365617263685F7061747465726E5F6D6174636865643D5E2828282872737C5253295B205D3F293F285C647B337D5B5C2D5C735D3F5C647B332C347D5B705061415D3F29297C283235285C647B387D7C5C647B317D5C2D5C647B377D29292924267365617263685F7061747465726E5F6F726465723D31267365617263685F73745F6E6F726D616C697365643D59267365617263685F726573706F6E73655F616374696F6E3D267365617263685F747970653D52535F53544F434B5F4E554D424552267365617263685F77696C645F63617264696E675F6D6F64653D4E4F4E45267365617263685F6B6579776F72643D3137342D38373338267365617263685F6B6579776F72645F6170703D31373438373338267365617263685F636F6E6669673D3026}{RS Online} & Other \\\hline
Part \#40 & watertight connector female & 1 & €~10,26/pc & €~10,26 & \href{https://benl.rs-online.com/web/p/industrial-circular-connectors/1748847/?relevancy-data=7365617263685F636173636164655F6F726465723D31267365617263685F696E746572666163655F6E616D653D4931384E525353746F636B4E756D626572267365617263685F6C616E67756167655F757365643D656E267365617263685F6D617463685F6D6F64653D6D61746368616C6C267365617263685F7061747465726E5F6D6174636865643D5E2828282872737C5253295B205D3F293F285C647B337D5B5C2D5C735D3F5C647B332C347D5B705061415D3F29297C283235285C647B387D7C5C647B317D5C2D5C647B377D29292924267365617263685F7061747465726E5F6F726465723D31267365617263685F73745F6E6F726D616C697365643D59267365617263685F726573706F6E73655F616374696F6E3D267365617263685F747970653D52535F53544F434B5F4E554D424552267365617263685F77696C645F63617264696E675F6D6F64653D4E4F4E45267365617263685F6B6579776F72643D3137342D38383437267365617263685F6B6579776F72645F6170703D31373438383437267365617263685F636F6E6669673D3026}{RS Online} & Other \\\hline
\end{longtabu}

\section{Build instructions}
\label{sec:Build_instructions}
OpenFish is designed for a rapid manufacturing and assembly process. The detailed build instructions are described below.

\begin{figure}[ht]
\centering%
\includegraphics[width=0.6\textwidth]{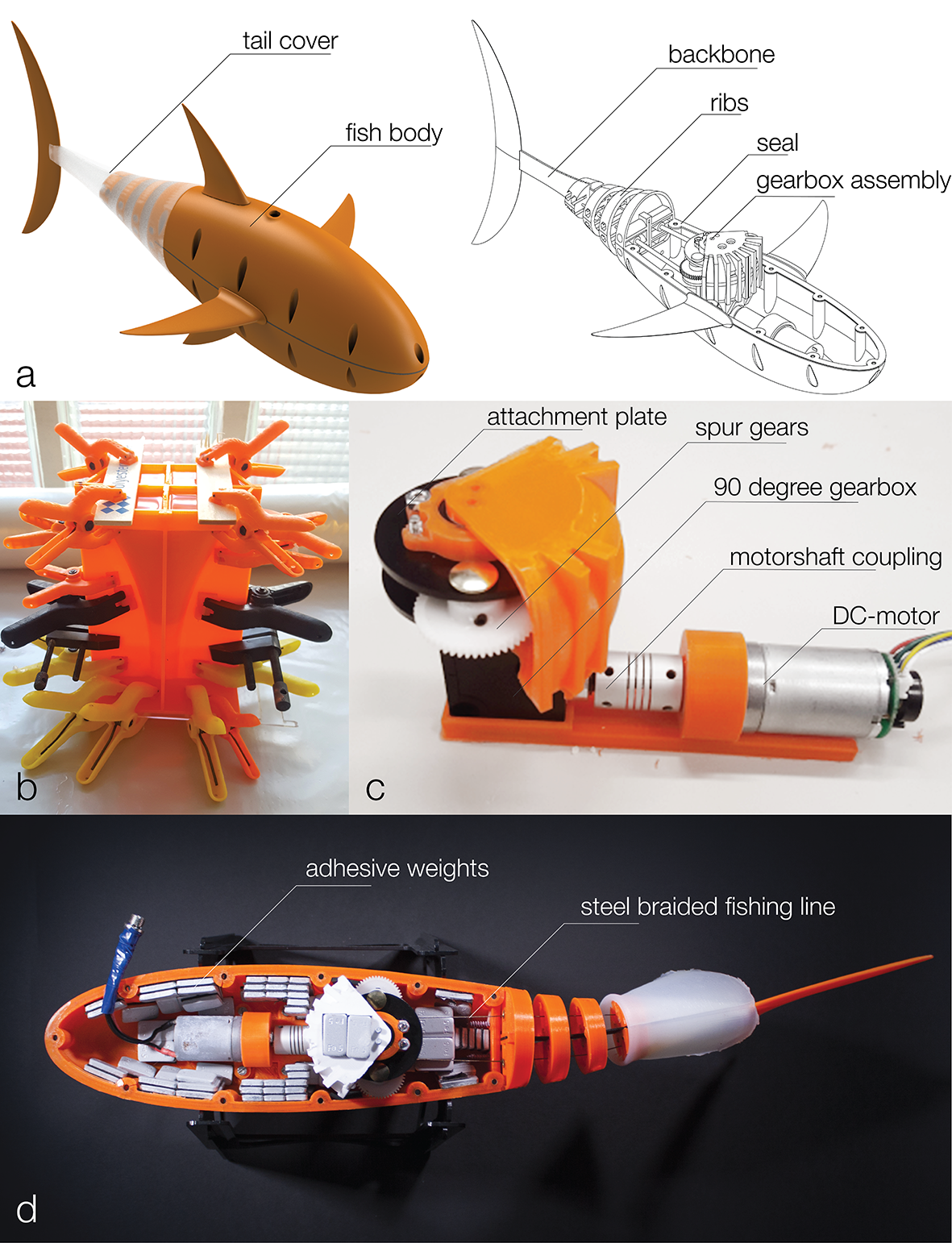}
\caption{Fabrication of the soft robotic fish. (a) CAD-model of the soft robotic fish with an indication of its main components, (b) casting of the silicone tail cover, (c) propulsion system with an indication of its main components, and (d) assembly of the soft robotic fish with an indication of the remainder components.}
\label{fig:fabrication}
\end{figure}

\begin{enumerate}
    \item Additive manufacturing of \textbf{Part~\#1-9}, \textbf{Part~\#11},\textbf{Part~\#12}, \textbf{Part~\#15-17} and \textbf{Part~\#19} (STL files are provided). All parts are fabricated using Fused Deposition Modeling (FDM) on an Ultimaker S5 using the Cura slicer engine. All parts are printed with PLA and an infill percentage of 100\%. The 3D printed body halves might not be fully waterproof directly after printing. In this case, the printing temperature can be slightly increased to improve the layer adhesion. In addition, a waterproof clear coat (e.g. an acrylic based clear spray coating) can be applied after printing.
    \item Lasercut \textbf{Part~\#13}, \textbf{Part~\#14}, \textbf{Part~\#18}, \textbf{Part~\#20-25} (DXF files are provided). The sheet thickness and materials for the various parts are indicated in the bill of materials summary. 

    \item Use \textbf{Part~\#11-13} to assemble the mold according to \textbf{Assembly~\#2}. Cast silicone in the mold to create \textbf{Part~\#10} (see Fig.~\ref{fig:fabrication}(b)). 
    \item Assemble the DC-motor driven propulsion system according to \textbf{Assembly~\#1} (see Fig.~\ref{fig:fabrication}(c)). To implement the optional steering system, include \textbf{Part~\#30} as indicated in \textbf{Assembly~\#1} and include a servo motor to rotate the spur gear and enable steering. In case the steering system is not required, the two spacer bolts (\textbf{Part \#33}) that are positioned furthest away from the tail should be placed upside down with respect to their position in \textbf{Assembly~\#1}. Then, two bolts can be secured to connect the two spacer bolts to \textbf{Part~\#17} and fix the steering system.
    \item Assemble the complete fish according to \textbf{Assembly~\#1} (see Fig.~\ref{fig:fabrication}(d)). The cables (\textbf{Part~\#38}) are pulled tight and fastened with in the pushrod connectors. It is important to ensure that the length of the cable is the same on both sides. The total volume of OpenFish is approximately $1370~\mathrm{cm^3}$ (or $1340~\mathrm{cm^3}$ excluding the dorsal, anal, and pectoral fins). The body of the fish is weighed and its weight is subtracted from the target weight of 1370~g. The missing weight minus 15 grams is added by placing adhesive weights (\textbf{Part~\#37}) on the inside of the bottom shell. Then, the fish is closed by putting the rubber seal (\textbf{Part~\#14}) between the two halves of the fish (\textbf{Part~\#1} and \textbf{Part~\#2}). and tightening the bolts. Adhesive weights are now added on the outside of the fish until neutral buoyancy is achieved. Next, the fish is opened up and the adhesive weights that were placed on the outside of the fish are placed on the inside on the same frontal plane. After closing the fish again, waterproof tape is applied on the connection between the silicone tail cover (\textbf{Part~\#10}) and the main body of the fish to ensure a waterproof seal.

\end{enumerate}

\section{Operation instructions}
\label{sec:Operation_instructions}

\begin{figure}[ht]
\centering%
\includegraphics[width=0.7\textwidth]{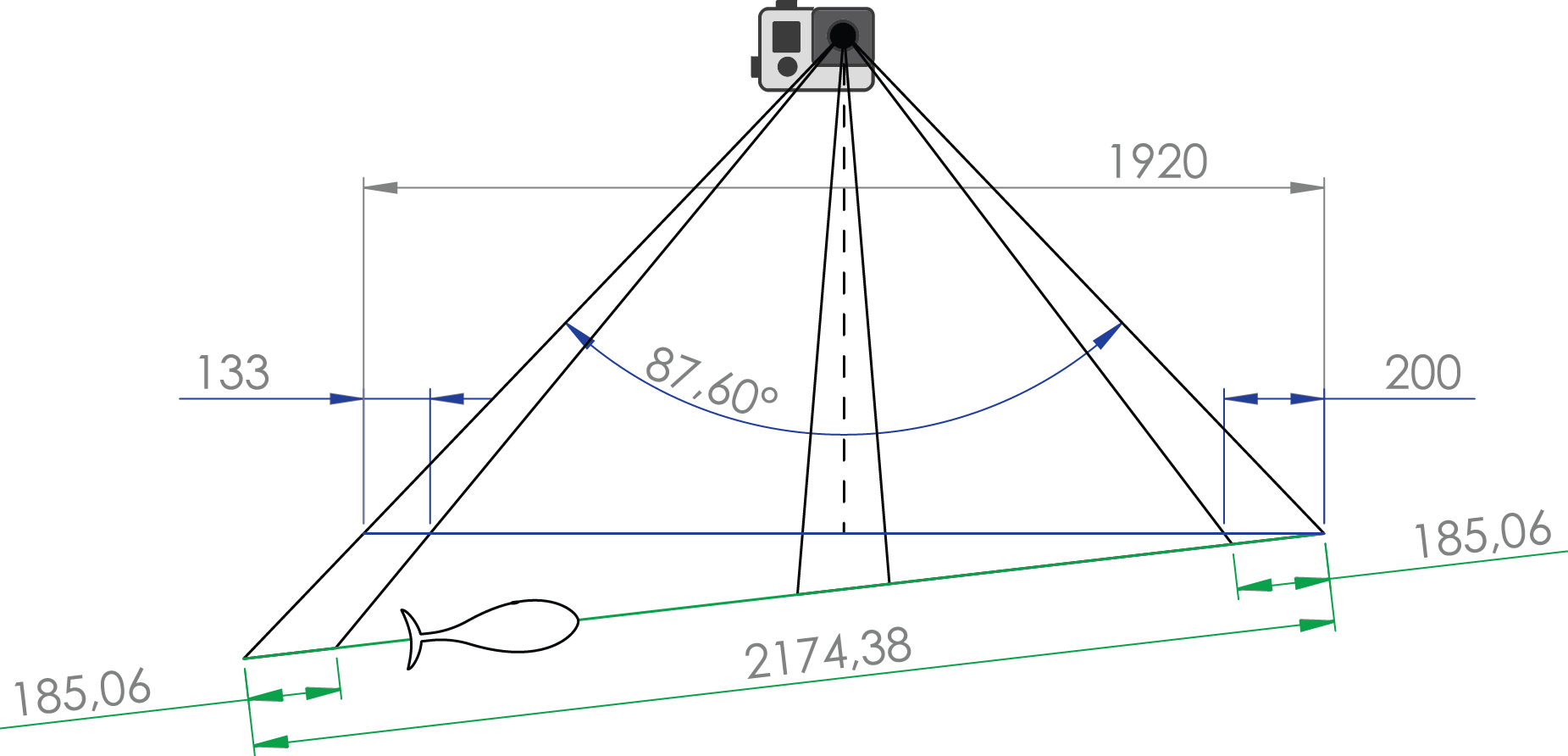}4\caption{Example calculation of the traveled distance. The camera makes an angle of $87.6$ degrees. A segment with a known length of $0.2481~\mathrm{m}$ is $133$px long in the initial position, and $200~\mathrm{px}$ long at the end of the swimming interval. The traveled distance over the interval can thus be calculated as $2174.38~\mathrm{px}$.}
\label{fig:setup}
\end{figure}

This section describes the setup and protocol that was used to operate the soft robotic fish and to measure its performance. The setup can be easily reproduced in order for others to benchmark new versions of the open source soft robotic fish. It is important to note that the fish is free-swimming (the head is not fixed), such that head sway and other stability issues will influence the swimming performance. Therefore, the OpenFish design can also be used to study these phenomena. The basin is sufficiently wide to exclude effects from the walls of the basin on the flow, and long enough to allow the fish to reach its top speed. The robotic fish is tethered to allow for reliable communication. A camera (GoPro7, $1080~\mathrm{p}$ at $120~\mathrm{fps}$) is placed above the water with linear settings to minimize distortion due to the lens. Fig.~\ref{fig:setup} shows how the swimming speed $U$ is calculated using a segment of the fish with a known length for calibration of the traveled distance. Therefore, it is not necessary for the fish to swim perfectly parallel to the water during the interval of interest. In this work, the performance of the fish at different tail beat frequencies was studied. The voltage applied to the motor was incremented with steps of $0.5~\mathrm{V}$, starting from a minimum voltage $5~\mathrm{V}$ and until failure of the prototype or until the maximum motor specifications were reached. For each voltage, three straight swimming intervals were captured. The tail beat frequency $f$ was measured from the video footage. The tail sweep amplitude $A$ is measured when the fish is underneath the camera in order to minimize camera distortion, and calibrated using the segment with known length. Using these parameters, the Strouhal number can be calculated as:

\begin{equation}
    St=\dfrac{f*A}{U}
\end{equation}
The Strouhal number is a dimensionless quantity that is used for the analysis of unsteady fluid flow dynamics problems.  Experiments have demonstrated high peak propulsive efficiency for intermediate Strouhal numbers (between $0.2$ and $0.4$)~\cite{Taylor2003}. This peak in propulsive efficiency is associated with maximum amplification of the shed vortices in the wake. Therefore, Strouhal numbers between $0.2$ and $0.4$ are an indicator for good swimming efficiency.

\section{Validation and characterization}
\label{sec:Validation_and_characterization}

\begin{figure}[ht]
\centering%
\includegraphics[width=0.45\textwidth]{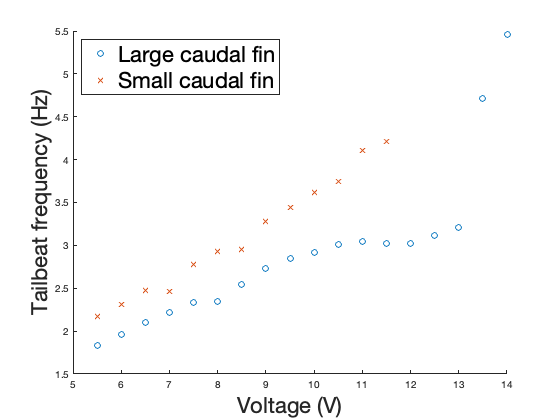}
\includegraphics[width=0.45\textwidth]{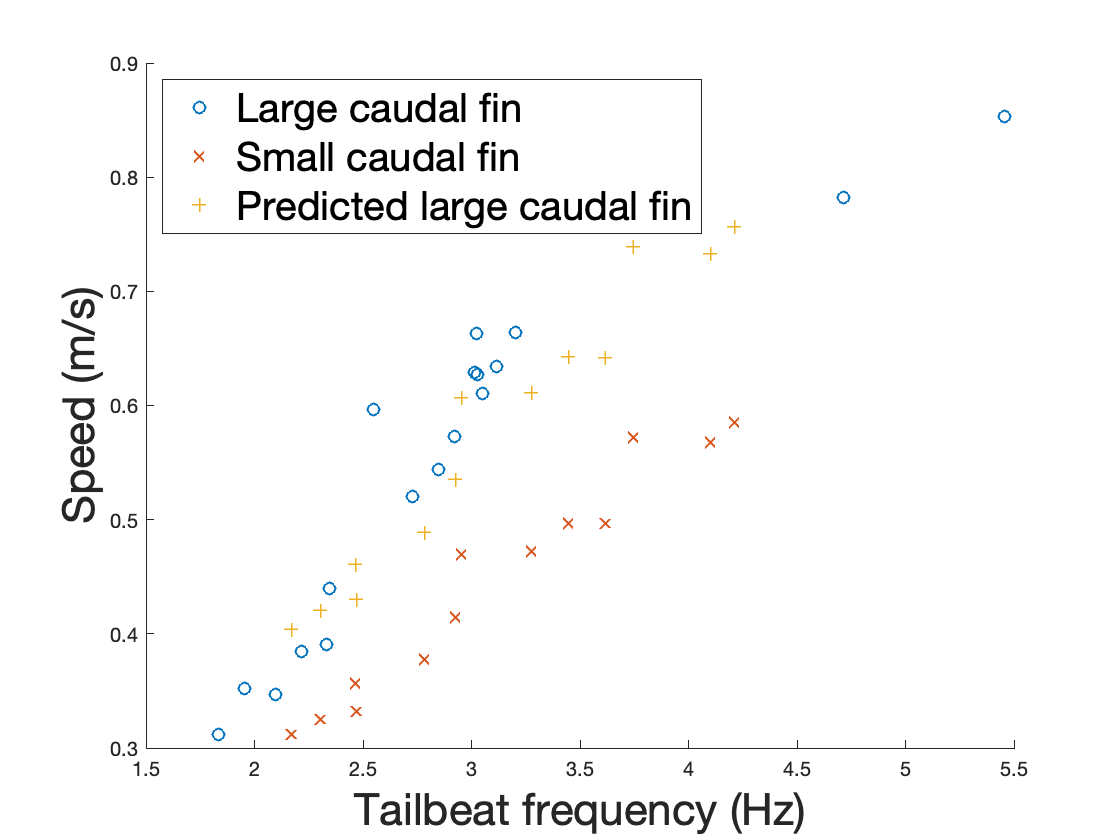}
\caption{The tail beat frequency over voltage (left), and swimming speed over tail beat frequency (right) for both caudal fin designs (see Fig.~\ref{fig:design}(b)). The yellow points indicate the predicted speed of the fish with the large caudal fin based on the speed achieved with the small caudal fin and the surface areas of both caudal fins, using eq.~(\ref{eq:predictedspeed}).}
\label{fig:frequencyspeed}
\end{figure}

This section briefly discusses the performance of the open source soft robotic fish. A more detailed evaluation of the performance of the soft robotic fish is given in our previous work~\cite{vandenBerg2020}. The performance of the robotic fish is also demonstrated in the supplementary video (\href{https://youtu.be/tvL4VXgySOI}{https://youtu.be/tvL4VXgySOI}). It can be seen that the fish accurately reproduces a thunniform-like swimming mode. The soft robotic fish was able to achieve a top speed of $0.85~\mathrm{m/s}$ ($2.02$ bodylengths per second ($\mathrm{BL/s}$)) at a tail beat frequency of $5.46~\mathrm{Hz}$. For tail beat frequencies above $2.33~\mathrm{Hz}$, the Strouhal number drops between $0.31$ and $0.4$. This is a significant improvement over the servo-driven soft robotic fish by Zhong et al., which achieved Strouhal numbers between $0.36$ and $0.6$.~\cite{zhong2017novel}. The tailbeat frequency over voltage and the swimming speed over tail beat frequency for both caudal fin designs is shown in Fig.~\ref{fig:frequencyspeed}. The speeds that were achieved with the larger caudal fin are significantly higher than those achieved with the smaller caudal fin at the same tail beat frequencies. Figure.~\ref{fig:frequencyspeed} also shows the predicted speed for the large caudal fin based on the speed that was achieved with the small caudal fin and the surface areas of both caudal fins, using eq.~(\ref{eq:predictedspeed}). These predictions are in good agreement with the actual speed obtained with the large caudal fin. The performance of the large tail starts to exceed the predictions from a tailbeat frequency of approximately $2.5~\mathrm{Hz}$. The footage shows that around this frequency the tail starts to form an S-shape due to larger deformation of the passive tail segment (as shown in Fig.~\ref{fig:fishlocomotion}(d)). This is associated with higher efficiencies as the caudal fin has a more optimal angle of attack.\\

\noindent
\textbf{CRediT author statement}\\
\noindent

\noindent
\textbf{Sander van den Berg}: Conceptualization, Methodology, Investigation, Writing- Original draft preparation, Visualization \\ 
\textbf{Rob Scharff}: Conceptualization, Methodology, Supervision, Writing- Original draft preparation, Visualization\\ \textbf{Zolt{\'a}n Rus{\'a}k}: Conceptualization, Methodology, Supervision, Writing- Reviewing and Editing \\ \textbf{Jun Wu}: Conceptualization, Methodology, Supervision, Writing- Reviewing and Editing \\

\noindent
\textbf{Acknowledgements}\\
This research did not receive any specific grant from funding agencies in the public, commercial, or not-for-profit sectors.

\bibliographystyle{ieeetr}
\bibliography{References_clean.bib}

\end{document}